\documentclass[runningheads]{llncs}
\usepackage[T1]{fontenc}
\usepackage{graphicx}
\usepackage{todonotes}
\usepackage{latexsym}
\usepackage{comment}
\usepackage{graphicx}
\usepackage{enumitem}
\usepackage{hyperref}
\usepackage{color}
\usepackage{booktabs} 
\usepackage{multirow}
\usepackage{amssymb}
\usepackage{xcolor}

\usepackage[normalem]{ulem}
\usepackage{soul}
\definecolor{ashgrey}{rgb}{0.8, 0.8, 0.8}

\usepackage{placeins}

\usepackage{fontawesome5}
\usepackage{xcolor,colortbl}

\definecolor{ashgrey}{rgb}{0.8, 0.8, 0.8}
\usepackage{amsmath}
\usepackage{amssymb}
\usepackage{latexsym}
\usepackage{comment}
\usepackage{graphicx}
\usepackage{enumitem}
\usepackage{hyperref}
\usepackage{color}
\usepackage{booktabs} 
\usepackage{multirow}
\usepackage{amssymb}
\usepackage{xcolor}
\usepackage{amsmath}
\usepackage{placeins}
\usepackage{fontawesome5}
\usepackage{xcolor,colortbl}
\usepackage{soul}
\usepackage{xcolor}
\definecolor{ashgrey}{rgb}{0.8, 0.8, 0.8}
\usepackage{amsmath}
\usepackage{amssymb}
\usepackage[misc]{ifsym}

\newcommand\redsout{\bgroup\markoverwith{\textcolor{red}{\rule[0.5ex]{2pt}{0.4pt}}}\ULon}

\definecolor{darkturquoise}{rgb}{0.0, 0.81, 0.82}
\definecolor{lightblue}{rgb}{.50,.95,1}
\definecolor{tri}{rgb}{.25,.88,.82}
\definecolor{lilac}{rgb}{0.85,0.64,0.85}

\definecolor{lightblue}{rgb}{0.53, 0.81, 0.98}

\begin{document}
\title{Propaganda to Hate: A Multimodal Analysis of Arabic Memes with Multi-Agent LLMs}
\titlerunning{Propaganda to Hate: A Multimodal Analysis}
%
\author{%
Firoj Alam\inst{1}\textsuperscript{\Letter}\orcidID{0000-0001-7172-1997} \and
Md. Rafiul Biswas\inst{2}\orcidID{0000-0002-5145-1990} \and
Uzair Shah\inst{2}\orcidID{0000-0002-6729-5654} \and
Wajdi Zaghouani\inst{3}\orcidID{0000-0003-1521-5568} \and
Georgios Mikros\inst{2}\orcidID{0000-0002-4093-5973}
}
\authorrunning{Alam et al.}
%
\institute{%
$^1$Qatar Computing Research Institute, Qatar, 
$^2$Hamad bin Khalifa University, Qatar, \\
$^3$Northwestern University in Qatar, Qatar \\
fialam@hbku.edu.qa\\
\textsuperscript{(\Letter)}Corresponding author
}

\maketitle              
\begin{abstract}
In the past decade, social media platforms have been used for information dissemination and consumption. While a major portion of the content is posted to promote citizen journalism and public awareness, some content is posted to mislead users. Among different content types such as text, images, and videos, memes (text overlaid on images) are particularly prevalent and can serve as powerful vehicles for propaganda, hate, and humor. In the current literature, there have been efforts to individually detect such content in memes. However, the study of their \textit{intersection} is very limited. In this study, we explore the \textit{intersection} between propaganda and hate in memes using a multi-agent LLM-based approach. We extend the propagandistic meme dataset with coarse and fine-grained hate labels. Our finding suggests that there is an association between propaganda and hate in memes. We provide detailed experimental results that can serve as a baseline for future studies. We will make the experimental resources publicly available to the community.\footnote{\url{https://github.com/firojalam/propaganda-and-hateful-memes.git}}  
\end{abstract}


\section{Introduction} 
\label{sec:introduction}

Social media has emerged as a primary channel for freely sharing content online. Its exponential growth has significantly transformed the landscape of information dissemination. However, misuse of these platforms has made them fertile grounds for the spread of inappropriate content, misinformation, and disinformation~\cite{alam-etal-2022-survey}. While interactions on social media facilitate public discussions on a range of topics, from local issues to politics, they also harbor and propagate hate speech and offensive content through various content types, text, images, and videos \cite{mubarak2023detecting,fortuna2018survey,ijcai2022p781,alam-etal-2022-survey}.
 
To address such problems across different modalities, there have been efforts to automatically detect them using both mono- and multimodal modeling approaches \cite{da2019fine}. For propagandistic content detection, research efforts have specifically focused on defining techniques and tackling the issue across various types of content, including news articles, tweets, memes, and textual content in multiple languages \cite{dimitrov2021detecting,propaganda-detection:WANLP2022-overview,piskorski-etal-2023-semeval}. Similarly, significant efforts have been made in hate-speech detection \cite{schmidt-wiegand-2017-survey,davidson2017automated}. A notable initiative in meme research is the Hateful Memes Challenge \cite{kiela2020hateful}, which has inspired many subsequent studies.  

Our research lies at the intersection of multimodal content analysis, propaganda detection, and hate speech identification. While progress has been made in these fields for English and other high-resource languages, the research for Arabic  remains underexplored.
Building on the work of \cite{alam2022overview} and \cite{hasanain2023large} in Arabic propaganda detection, our study analyzes Arabic memes, addressing a gap in the literature. The findings can assist social media platforms, policymakers, and civil society organizations in combating harmful online content.

The key aspects of our work are as follows: \textit{(i)} we present a novel multi-agent LLM-based approach to analyze the association between propaganda and hateful memes in Arabic social media content; \textit{(ii)} we demonstrate the application of multi-agent LLM systems for automated annotation of complex, multimodal data, offering a scalable solution for processing large volumes of data in low-resource settings; \textit{(iii)} in addition to coarse-grained hateful categories, we also explore fine-grained categorization of hateful and non-hateful memes; \textit{(iv)} we provide experimental results on both coarse and fine-grained categories; \textit{(v)} finally, we will make the dataset of hateful memes available to the community.

\section{Related Work}
\label{sec:related_work}

\subsection{Propagandistic Content}
\textbf{Textual Content:} The study of propagandistic content has attracted significant attention in recent years. 
Da et al. (2020) introduced a large-scale dataset for fine-grained propaganda detection in news articles, presenting a corpus of 350K sentences annotated with 18 propaganda techniques \cite{da2020semeval}. The annotation schema has been extended to include 23 techniques and a multilingual corpus has been proposed in \cite{piskorski-etal-2023-semeval}. Following the same annotation schema, datasets have been developed for Arabic and shared tasks has been organized \cite{propaganda-detection:WANLP2022-overview,hasanain-etal-2023-araieval,hasanain2024can}.


\noindent
\textbf{Multimodal Content:} Building on previous research in textual content, Dimitrov et al. (2021) introduced SemEval-2021 Task 6, which focuses on detecting persuasion techniques in texts and images within memes \cite{dimitrov2021semeval}. Subsequently, the focus has expanded to include the detection of multilingual and multimodal propagandistic memes \cite{dimitrov2024semeval}. Similarly, related work on Arabic involves the development of datasets and a shared task for propaganda detection \cite{alam2024armeme,araieval:arabicnlp2024-overview}.
Fang et al. (2022) used separate networks to embed text and images, fusing these multi-modal embeddings. A split-and-share module with multi-level representations was employed to improve persuasive technique detection \cite{fang2022emotion}. 

\subsection{Hateful Memes}
The study of hateful memes presents unique challenges due to their multimodal nature. Kiela et al. (2020) introduced the Hateful Memes Challenge, a large-scale dataset and benchmark for multimodal hate detection. This work highlighted the importance of integrating both textual and visual elements to identify hate speech in memes \cite{kiela2020hateful}.
Addressing the challenge of low-resource languages, datasets have also been developed in various languages \cite{hossain2024deciphering}. In \cite{ijcai2022p781}, the authors provide a detailed survey of multimodal and harmful memes, highlighting the significance of the problem and proposing future research avenues. 


\subsection{Multi-Agent Systems in Content Analysis}
The application of multi-agent systems to content analysis is an emerging field, which could be an effective approach in analyzing complex narratives across various media~\cite{guo2024large}. 
Chen et al. (2021) introduced a dynamic content moderation system using multi-agent reinforcement learning, which adapts based on user interactions and content patterns for improved detection of harmful content \cite{chen2021multi}. 
These studies emphasize the value of multi-agent systems in analyzing complex, often propagandistic or hateful content in memes.
Building on this, our work focuses on Arabic memes, employing a multi-agent LLM approach. 
We explore the association between propaganda and hateful memes in low-resource settings. We employ LLMs as multiple agents to automate the data annotation process, demonstrating the utility of LLMs as data annotators in detecting hateful memes.




\section{Dataset}
\label{sec:dataset}

\subsection{Propagandistic Memes}
For this study, we used ArAIEval-2024 dataset \cite{araieval:arabicnlp2024-overview}, which consists of approximately $\sim$3k memes, each annotated with labels as either propagandistic or not-propagandistic. These memes were collected from various social media platforms, including Facebook, Twitter, Instagram, and Pinterest. We have annotated each meme by three annotators, with the final label determined by majority vote.
The text from the memes was extracted using an off-the-shelf OCR tool\footnote{\url{https://github.com/JaidedAI/EasyOCR}}, followed by manual corrections for propagandistic memes. The distribution of propagandistic and not-propagandistic labels is 40\% and 60\%, respectively. Further details about this dataset are available in \cite{araieval:arabicnlp2024-overview}, and the comprehensive annotation guidelines are provided in \cite{alam2024armeme}.
For the experiments, the dataset is split into 70\%, 10\%, and 20\% for training, development, and testing, respectively. 


\begin{figure}[t]
    \centering
    \includegraphics[scale=0.22]{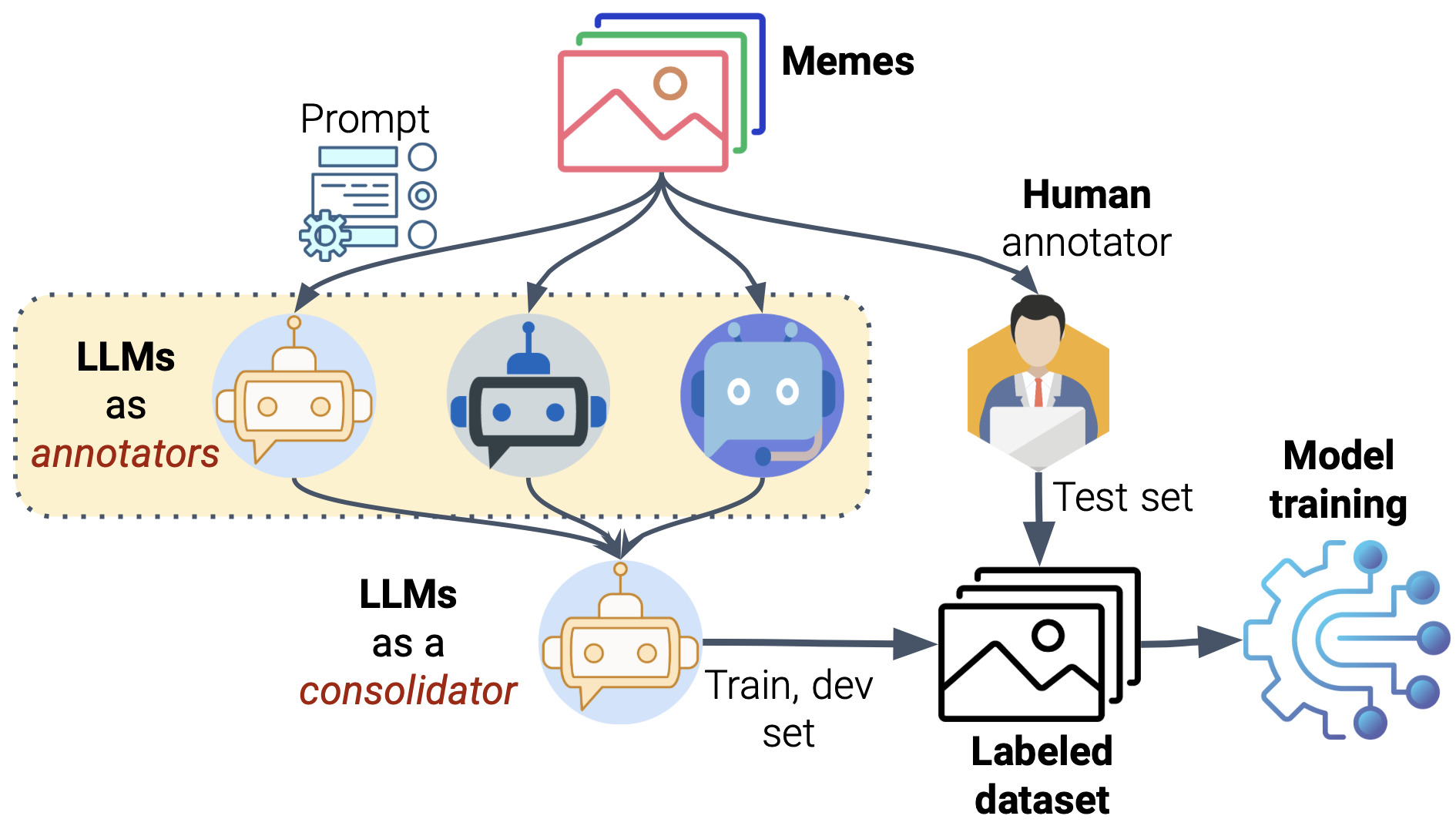}
    \caption{Experimental pipeline with LLM agents as annotators and consolidator.}
    \label{fig:experiment_setup}
    \vspace*{-0.3cm}
\end{figure}

\subsection{Hatefulness and Fine-grained Categories}
For the hatefulness and their fine-grained categorization we used ArAIEval-2024 dataset, mentioned earlier. Our motivation to use ArAIEval-2024 dataset is that this is the only meme dataset currently available for Arabic, which has already been annotated for propagandistic content. Another motivation was to understand the association between propagandistic and hateful memes. In Figure \ref{fig:experiment_setup}, we provide full pipeline for the data preparation to classification experiments.

\subsubsection{LLM Agents as Annotators}
\label{sec-llm-agents}
To employ LLM agents as annotators, we selected three well-known and top-performing commercial models: OpenAI's GPT-4o \cite{openai2023gpt}, Google's Gemini Pro (version 1.5) \cite{team2023gemini}, and Claude 3.5 (Sonnet).\footnote{\url{https://www.anthropic.com/news/claude-3-5-sonnet}}.  
For the annotation process, we use the same manual procedure discussed in \cite{hasanain2023large}, which involves a two-phase approach. In the first phase, known as the \textit{annotation phase}, three annotators independently annotate memes following the guidelines outlined in \ref{ssec-manual-annotation}. In the second phase, known as the \textit{consolidation phase}, we review and resolve any disagreements from the annotations received during the first phase. As illustrated in the figure, highlighted in dark red, we employ LLM agents as \textit{annotators} in the first phase and as a \textit{consolidator} in the second phase. For each phase, we use a specific prompt in a zero-shot setup for the LLM agent. Following the annotation guidelines discussed below, we ask an LLM agent to perform two tasks: \textit{(Task 1)} label each meme as hateful or not-hateful, and \textit{(Task 2)} based on the label from Task 1, provide a fine-grained categorization. For example, if a meme is categorized as hateful in Task 1, it should then provide a fine-grained label from one of the eight categories mentioned below. 
The prompt in the second phase is slightly different. Here, the task also involves considering the labels obtained from the first phase to make a final decision. For this phase, we have experimented with using GPT-4o as the consolidator. 

We used this LLM-based multi-agent approach for the training and development (dev) sets. To validate the quality of the multi-agent approach, we quantified the labels provided by each LLM agent by comparing them with human-annotated labels on the test dataset.

\subsubsection{Manual Annotation}
\label{ssec-manual-annotation}
To verify the LLM-based multi-agent approach, we manually annotated a test set from the ArAIEval-2024 dataset, as shown in Figure \ref{fig:experiment_setup}. For the annotation process, we developed a set of instructions, which are discussed below. The typical approach to annotation involves two to three annotators. However, for this study, we relied on a single annotator who had prior experience with similar annotation tasks. 

\subsubsection{Annotation Instructions}
\label{ssec-manual-annotation-instructions}

The purpose of this annotation is to identify whether a meme is hateful or not-hateful. A hateful meme can attack different individuals, organizations or entities. Therefore, another task is identifying the attack types. A non-hateful meme can be humorous or sarcastic. Therefore, the idea to also identify the sub-categories within non-hateful memes. We adopted the annotation definition and instructions from prior work \cite{kiela2020hateful,mathias-etal-2021-findings}. Below we provide the definitions: 

\paragraph{\textbf{Hateful:}} A direct or indirect attack on individuals based on characteristics such as ethnicity, race, nationality, immigration status, religion, caste, sex, gender identity, sexual orientation, disability, or disease. We define an attack as hate-speech that is violent, dehumanizing (such as comparing individuals to non-human entities like animals), involves statements of inferiority, or calls for exclusion or segregation. Mocking hate crimes is also classified as hate speech. However, attacks directed at groups that perpetuate hate (e.g., terrorist organizations) are not considered hate speech.\footnote{\url{https://transparency.meta.com/en-gb/policies/community-standards/hate-speech/}}

\textbf{Fine-grained categories hatefulness:} 
\begin{itemize}[noitemsep,topsep=0pt,leftmargin=*,labelwidth=!,labelsep=.5em]
    \item \textbf{Dehumanizing:} Explicitly or implicitly portraying or describing a group as subhuman.
    \item \textbf{Inferiority:} Asserting that a group is inferior, less worthy, or less important than society as a whole or compared to another group.
    \item \textbf{Inciting violence:} Explicitly or implicitly advocating for harm to be inflicted on a group, including physical violence.
    \item \textbf{Mocking:} Joking about, ridiculing, demeaning, or disparaging a group.
    \item \textbf{Contempt:} Expressing strong negative emotions or feelings toward a group.
    \item \textbf{Slurs:} Using biased or derogatory terms to refer to, describe, or characterize a group.
    \item \textbf{Exclusion:} Advocating for, planning, or justifying the exclusion or segregation of a group from society as a whole or from specific areas.
    \item \textbf{Other:} None of the above.
\end{itemize}

\paragraph{\textbf{Not-Hateful:}}
The content is humorous, neutral, or positive, without targeting or harming specific individuals or groups. It is light-hearted and intended for entertainment without being offensive. Additionally, the content does not promote or incite violence, hatred, or discrimination.

\textbf{Fine-grained not-hateful categories:}
\begin{itemize}[noitemsep,topsep=0pt,leftmargin=*,labelwidth=!,labelsep=.5em]
    \item \textbf{Humor:} The purpose of humor is to entertain, amuse, or bring joy to the audience. Often characterized by jokes, puns, or playful language. Humor can vary widely in style, including wit, slapstick, parody, and satire.
    \item \textbf{Sarcasm:}  Typically involves saying the opposite of what one means. Sarcasm is a form of irony that always occurs with a deliberate mismatch between what is said and what is meant, intentionally to ridicule or mock a specific target.
    \item \textbf{Other:} None of the above
\end{itemize}

\subsubsection{Annotated Dataset}

As discussed in Section \ref{sec-llm-agents}, we annotated the test data into `Hateful' and `Not-Hateful' categories using GPT-4o, Sonnet (Claude 3.5), and Gemini (Vertex). We then provided the three annotated labels (obtained from GPT-4o, Sonnet, and Gemini) as prompts to GPT-4o and asked it to choose the best label that matches the data. The generated output label is termed \textit{GPT-4o consolidation}. Table \ref{tab:annotation-agreement} shows the inter-annotator agreement (IAA) among the annotators. We computed the annotation agreement using pairwise Cohen's kappa score in different setups: \textit{(i)} LLMs as annotators \textit{vs.} an LLM as a consolidator, \textit{(ii)} LLMs as annotators \textit{vs.} human annotation, and \textit{(iii)} pairwise between LLMs as annotators.

It shows that the IAA between \textit{GPT-4o} and \textit{GPT-4o consolidation} is high (0.786), representing substantial agreement. It is reasonable because the \textit{GPT-4o consolidation} is derived from the GPT-4o, which means that the consolidated label inherently aligns closely with the labels initially provided by GPT-4o. Interestingly, we observe that the IAA between Sonnet and \textit{GPT-4o consolidation} is significant and high (0.701), which denotes Sonnet's capability to understand hateful content and memes. 

Table \ref{tab:annotation-agreement} shows that the IAA between Sonnet and the human annotator achieved a higher score (0.405) compared to other annotation labels. We also performed pairwise agreements among LLM annotators and found that the agreement between Sonnet and GPT-4o is higher (0.528). Finally, we measured the annotation agreement among all three annotators 
and obtained a score of 0.369. 

The annotation agreements between the three different LLMs and the human annotator suggest that Sonnet has a fair capability of understanding `Hateful' content compared to a human annotator. Therefore, we used Sonnet to annotate the training and development datasets. 

\begin{table}[]
\centering
\caption{Annotation agreement for different setups.}
\label{tab:annotation-agreement}
\setlength{\tabcolsep}{2pt}
\scalebox{0.8}{%
\begin{tabular}{@{}llr|llr@{}}
\toprule
\multicolumn{1}{c}{\textbf{Anno. 1}} & \multicolumn{1}{c}{\textbf{Anno. 2}} & \multicolumn{1}{c|}{\textbf{Kappa}} & \multicolumn{1}{c}{\textbf{Anno. 1}} & \multicolumn{1}{c}{\textbf{Anno. 2}} & \multicolumn{1}{c}{\textbf{Kappa}} \\ \midrule
\multicolumn{3}{c|}{\textbf{Agreement: LLMs vs. LLM as a Consolidator}} & \multicolumn{3}{c}{\textbf{Agreement: LLMs vs Human}} \\ \midrule
GPT-4o & GPT-4o & 0.786 & GPT-4o & Human & 0.233 \\
Sonnet (Claude) & GPT-4o & 0.701 & Claude-3.5 (Sonnet) & Human & 0.405 \\
Gemini-1.5 (Vertex) & GPT-4o & 0.236 & Gemini-1.5 (Vertex) & Human & 0.300 \\\cmidrule{1-3}
\multicolumn{3}{c|}{\textbf{Agreement: LLMs (Pairwise)}} & GPO-4o Consolidation & Human & 0.300 \\  \cmidrule{1-3}
Gemini-1.5 (Vertex) & Claude-3.5 (Sonnet) & 0.266 &  &  & \multicolumn{1}{l}{} \\
GPT-4o & Gemini-1.5 (Vertex) & 0.142 &  &  & \multicolumn{1}{l}{} \\
Claude-3.5 (Sonnet) & GPT-4o & 0.528 &  &  & \multicolumn{1}{l}{} \\ \bottomrule
\end{tabular}
}
\end{table}

\textbf{Data Stat: Hateful Meme} Table \ref{tab:annotated-table} presents the distribution of class labels for training, development, and testing datasets, categorized into ``Hate/Not-hate'' and further labeled into fine-grained categories. 
There is a significantly larger number of ``Not-Hateful'' (N=1931) category instances compared to the ``Hateful'' (N=212) category. In the fine-grained label, ``Mocking'' has a notable presence (N=133) in the `Hateful' category. Similarly, in the fine-grained ``Not-Hateful'' categories, ``Humor'' (N=1815) overwhelmingly dominates, followed by ``Sarcasm''. This distribution highlights the imbalance in the data.


\begin{table}[h]
\centering
\caption{Distribution of annotated data: The training and development sets were labeled using Sonnet, while the test set was labeled by a human.}
\label{tab:annotated-table}
\setlength{\tabcolsep}{2pt}
\scalebox{0.8}{%
\begin{tabular}{@{}lrrr|lrrr@{}}
\toprule
\multicolumn{4}{c|}{\textbf{Hate/Not-hate}} & \multicolumn{4}{c}{\textbf{Hate: Fine-grained categories}} \\ \midrule
\multicolumn{1}{c}{\textbf{Label}} & \multicolumn{1}{c}{\textbf{Train}} & \multicolumn{1}{c}{\textbf{Dev}} & \multicolumn{1}{c|}{\textbf{Test}} & \multicolumn{1}{c}{\textbf{Label}} & \multicolumn{1}{c}{\textbf{Train}} & \multicolumn{1}{c}{\textbf{Dev}} & \multicolumn{1}{c}{\textbf{Test}} \\\midrule
Hateful & 212 & 32 & 154 & Contempt & 38 & 7 & 25 \\ 
Not-Hateful & 1,931 & 280 & 452 & Dehumanizing & 12 & 3 & 2 \\
\textbf{Total} & 2,143 & 312 & 606 & Mocking & 133 & 19 & 49 \\\cmidrule{1-4}
\multicolumn{4}{l}{\textbf{Non-Hate: Fine-grained categories}} & Inferiority & 5 & 1 & 14 \\  \cmidrule{1-4}
\multicolumn{1}{c}{Label} & \multicolumn{1}{c}{Train} & \multicolumn{1}{c}{Dev} & \multicolumn{1}{c}{Test} & Exclusion & \multicolumn{1}{c}{6} & \multicolumn{1}{c}{7} & \multicolumn{1}{c}{3} \\
Sarcasm & 105 & 19 & 118 & Inciting violence & 13 & 2 & 12 \\
Humor & 1,815 & 260 & 334 & Slurs & 6 & 1 & 29 \\
\textbf{Total} & 1,920 & 279 & 452 & Other & 10 & 1 & 20 \\
 & \multicolumn{1}{l}{} & \multicolumn{1}{l}{} & \multicolumn{1}{l}{} & \textbf{Total} & 223 & 41 & 154 \\ \bottomrule
\end{tabular}
}
\end{table}

\textbf{Propaganda and Hateful Meme}:
To understand the correlation between propaganda and hateful memes, we observe that out of 171 propagandistic memes in the test set, 56 memes are hateful (30\%) and 70\% are not hateful. This is possible because propagandistic memes may not always instigate hate or harm.

\section{Experiments}
\label{sec-experiments}
Our experiments consist of three setups: \textit{(i)} hate vs. not-hate, \textit{(ii)} fine-grained categories for hateful memes, and \textit{(iii)} fine-grained categories for non-hateful memes. These classification experiments involve unimodal (text and image) and multimodal classifications.


\textbf{Text classification}: We extracted text from propagandistic memes and applied various text classification techniques such as AraBERT, mBERT, CAMelBERT, and Qarib-BERT (\cite{antoun2020arabert,devlin-etal-2019-bert,abdelali2021pretraining}). The original dataset is imbalanced, and so we implemented a class weighting scheme during the fine-tuning process. Moreover, we optimized the model by adjusting the dropout rate. This approach led to significant improvements over using the original dataset alone. We embedded the LoRa to fine-tune the model 
in an efficient way that does not require fine-tuning all the parameters of the model. 
However, embedding LoRA 
did not improve the performance. We then fine-tuned 
AraBERT model.

\textbf{Image classification}: We 
fine-tuned 
ResNet50 and ConvNeXt-tiny \cite{liu2022convnet}. To ensure stable weight adjustments, we froze the feature extraction layers and fine-tuned
the classification layer. 
We also adjusted the dropout rate during the model training. 

\textbf{Multimodal classification}: 
To extract visual and text features, we applied ConvNext tiny and AraBERT models respectively. We combine the features using a fusion layer. We froze the visual models and trained the classification layer with textual data. We also used a dropout rate to improve the performance. 

\textbf{Experimental Setup}: 
We performed all of our experiments and trained our models on an Nvidia-RTX 2080 GPU. We employed the Adam optimizer with an initial learning rate of 1e-5 and 1e-4 for text and image, respectively. We used a batch size of 32, a sequence length of 128. 
We set the dropout rate to 0.25 for text data and trained for 50 epochs. For image data, we set a dropout rate of 0.5 and trained for 30 epochs. For multimodal data, we trained the model for 100 epochs with a stochastic drop rate of 0.2. Note that our choice of the models and parameters was inspired by prior studies \cite{alam2024armeme,shah-etal-2024-mememind}.


\begin{table}[]
\centering
\caption{Classification results on the test set for coarse and fine-grained hate labels across different modality setups.}
\label{tab:model-performance}
\setlength{\tabcolsep}{2pt}
\scalebox{0.8}{%
\begin{tabular}{@{}llrr|llrr@{}}
\toprule
\multicolumn{1}{c}{\textbf{Modality}} & \multicolumn{1}{c}{\textbf{Model}} & \multicolumn{1}{c}{\textbf{Acc}} & \multicolumn{1}{c|}{\textbf{M-F1}} & \multicolumn{1}{c}{\textbf{Modality}} & \multicolumn{1}{c}{\textbf{Model}} & \multicolumn{1}{c}{\textbf{Acc}} & \multicolumn{1}{c}{\textbf{M-F1}} \\ \midrule
\multicolumn{4}{c}{\textbf{Hate vs. Not-hate}} & \multicolumn{4}{c}{\textbf{Balanced dataset: Hate vs. Not-hate}} \\ \midrule
Text & AraBERT & 0.819 & 0.705 & Balanced & Fusion & 0.817 & 0.709 \\ \cmidrule(l){5-8}
Image & ConvNxT & 0.779 & 0.669 & \multicolumn{4}{c}{\textbf{Fine-grained label}} \\ \cmidrule(l){5-8}
Text+Image & Fusion & 0.764 & 0.709 & Hateful & Fusion & 0.224 & 0.166 \\
 &  &  &  & Not-Hateful & Fusion & 0.622 & 0.537 \\ \bottomrule
\end{tabular}
}
\end{table}

\section{Results and Discussion}
\label{sec:result_discussion}


In Table \ref{tab:model-performance}, we present the results for different classification setups. For hate vs. not-hate classification across different modalities, considering macro-F1, the text and multimodal models exhibit similar performance. For the multimodal model, we obtained model with a macro-F1 of 0.709 and an accuracy of 0.764. For the text modality, we obtained a macro-F1 of 0.705 and an accuracy of 0.818. For the image modality, we obtained a macro-F1 of 0.669 and an accuracy of 0.775.

To understand the class imbalance issue, we selected 500 propaganda labels and 500 non-propaganda labels from the dataset and applied the fusion model. This yielded a macro-F1 of 0.709 and an accuracy of 0.818. The hateful memes were further fine-grained into eight labels, while the non-hateful memes were fine-grained into two labels, as shown in Table \ref{tab:annotated-table}.
We applied the fusion model individually to the hateful and non-hateful fine-grained labels. The F1-score for the hateful fine-grained memes is 0.224, with an accuracy of 0.166, which is very low. This occurs because there are multiple labels within the hateful category, and the dataset is imbalanced for fine-grained hateful memes. In contrast, with only two labels in the non-hateful memes, the model performs better in classification, achieving a macro-F1 of 0.537 and an accuracy of 0.622. 

\section{Conclusion and Future Work}
\label{sec:conclusions}
In this study, we investigate whether the content in propagandistic memes may contain hate. To do so, we used a multi-agent LLM-based approach to label propagandistic memes with coarse and fine-grained hate categories. We observed that there is a moderate agreement between an LLM agent (Claude 3.5 Sonnet) and human annotation. This led us to label propagandistic memes with coarse and fine-grained hate categories. We further used the dataset to train the model and evaluate its performance on the test set. The developed dataset is skewed in nature, which also reflects its classification performance. It is important to note that this attempt can enable the development of a large-scale dataset in a cost-effective manner. The issue of label imbalance can be resolved with an increase in data size. Future study will investigate this direction further by increasing data size and exploring open-sourced LLMs.

\section{Acknowledgments}
\label{sec:ack}

The work of F. Alam, M. R. Biswas, U. Shah, W. Zaghouani and G. Mikros is partially supported by NPRP 14C-0916-210015 from the Qatar National Research Fund, part of Qatar Research Development and Innovation Council (QRDI).

%
%
\bibliographystyle{splncs04}
\bibliography{bib/main}

\end{document}